\documentclass[conference]{IEEEtran}
\IEEEoverridecommandlockouts

\usepackage{cite}
\usepackage{amsmath,amssymb,amsfonts}
\usepackage{algorithmic}
\usepackage{graphicx}
\usepackage{textcomp}
\usepackage{booktabs}
\usepackage{xcolor}
\usepackage{multirow}

\def\BibTeX{{\rm B\kern-.05em{\sc i\kern-.025em b}\kern-.08em
    T\kern-.1667em\lower.7ex\hbox{E}\kern-.125emX}}
\begin{document}

\title{Multi-Object Grounding via Hierarchical Contrastive Siamese Transformers\\
\thanks{ * Corresponding author}
}
\pdfoutput=1

\author{\IEEEauthorblockN{1\textsuperscript{st} Chengyi Du}
\IEEEauthorblockA{\textit{School of Information and Software Engineering } \\
\textit{University of Electronic Science and Technology of China}\\
Chengdu, China \\
duchengyi1224@gmail.com}
\and
\IEEEauthorblockN{2\textsuperscript{nd} Keyan Jin*}
\IEEEauthorblockA{\textit{University of Granada}\\
Granada, Spain \\
kjin@ieee.org}
}

\maketitle

\begin{abstract}
Multi-object grounding in 3D scenes involves localizing multiple objects based on natural language input. While previous work has primarily focused on single-object grounding, real-world scenarios often demand the localization of several objects. To tackle this challenge, we propose \textbf{H}ierarchical \textbf{Co}ntrastive \textbf{S}iamese \textbf{T}ransformers (H-COST), which employs a \textbf{Hierarchical Processing strategy} to progressively refine object localization, enhancing the understanding of complex language instructions. Additionally, we introduce a \textbf{Contrastive Siamese Transformer framework}, where two networks with the identical structure are used: one auxiliary network processes robust object relations from ground-truth labels to guide and enhance the second network, the reference network, which operates on segmented point-cloud data. This contrastive mechanism strengthens the model's semantic understanding and significantly enhances its ability to process complex point-cloud data. Our approach outperforms previous state-of-the-art methods by \textbf{9.5\% }on challenging multi-object grounding benchmarks.
\end{abstract}


\section{Introduction}
 The ability to effectively ground objects in 3D environments is essential for advancing various technological applications. 3D grounding serves as a foundational element for numerous tasks, including scene understanding\cite{peng2023openscene} and spatial reasoning\cite{vil3dref}, extending its impact on fields such as VR/AR, interactive embodied agents\cite{wang2023embodiedscan}, navigation\cite{saynav}, and robotic manipulation\cite{polaris}. Therefore, research has naturally progressed from 2D object grounding to the more complex challenge of 3D object grounding, which involves interpreting and localizing objects in three-dimensional space, often based on more ambiguous inputs like point-cloud data. 
 
 While significant progress has been made in single-object grounding \cite{vil3dref}, \cite{achlioptas2020referit3d}, \cite{yang2021sat} within the field of 3D object detection and grounding, multi-object grounding remains less explored despite its growing importance as real-world applications increasingly require simultaneous interaction with multiple objects. Recently, the task of multi-object 3D grounding\cite{zhang2023multi3drefer}  has been proposed. However, grounding multiple objects simultaneously introduces higher complexity, as it demands a more refined understanding of complex language instructions and a deeper semantic comprehension of the scene to accurately localize and describe objects in a cluttered 3D space.

\begin{figure}
    \centering
    \includegraphics[width=1.0\linewidth]{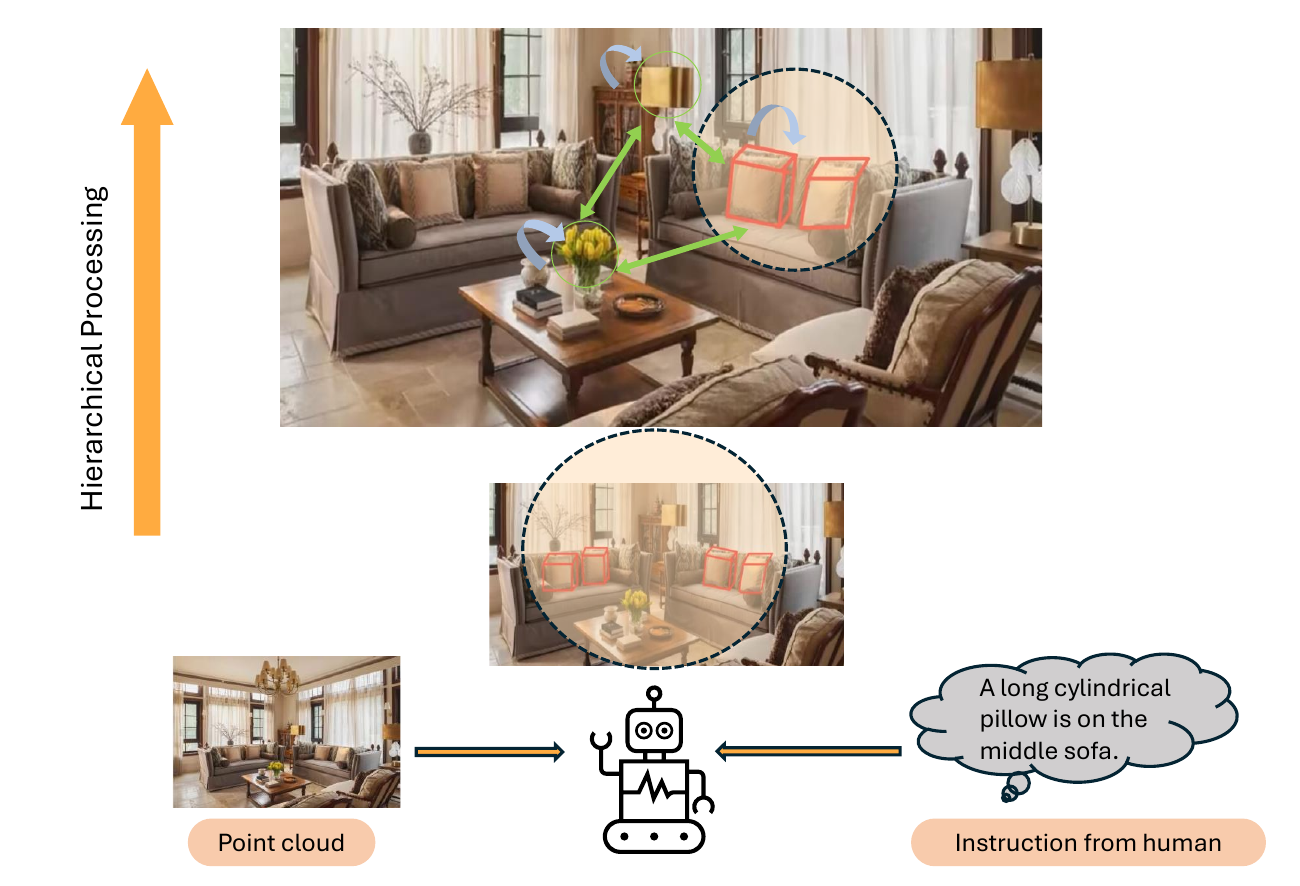}
    \caption{Hierarchical Contrastive Siamese Transformers (H-COST) starts with broad localization (bottom) to identify general regions and refines to precise localization . \textcolor{green}{Green arrows} show inter-object comparisons for differentiating objects, while \textcolor{blue}{blue arrows} indicate intra-object alignment for consistency. \textcolor{red}{Red bounding boxes} highlight objects refined through hierarchical processing. }
    \label{fig:teaser}
\end{figure}

       To address the challenges in multi-object grounding, as shown in Figure~\ref{fig:teaser} we propose \textbf{H}ierarchical \textbf{Co}ntrastive\textbf{ S}iamese \textbf{T}ransformers (H-COST), combining a hierarchical strategy with a contrastive learning framework. The hierarchical approach allows the model to initially identify potential targets broadly and then progressively refine their localization, mimicking the human-like process of object detection. improving its adaptability to scenes with varying object counts and enhancing its understanding of complex language instructions. 
       Recognizing that multi-object grounding often requires a step-by-step reasoning process, we draw inspiration from how humans typically approach such problems by gradually narrowing their focus. This iterative method allows the model to better handle the ambiguities and complexities inherent in spatial relationships, such as identifying objects that are next to, behind, or partially obscured by others. By emulating this human-like approach, our model achieves a more refined identification and localization of objects in cluttered environments, which contrasts with current methods that attempt to solve the problem in a single step. These single-step approaches often struggle with scenes where objects are densely packed or where subtle spatial relationships are critical to correct localization.

Building on this hierarchical process, we employ a Contrastive Siamese Transformer framework with two identical networks: an auxiliary network and a reference network. The auxiliary network captures robust object relations from ground-truth labels, focusing on extracting structured scene-level representations and relational dependencies. In contrast, the reference network operates on segmented point-cloud data, where the inherent sparsity and noise make it challenging to achieve accurate relational reasoning and semantic understanding. To enhance the inference capabilities of the reference network, the framework employs a contrastive mechanism that aligns the intermediate layer outputs of the two networks. By minimizing the disparity between their feature representations, the reference network progressively learns to approximate the robust relational understanding encoded in the auxiliary network. This alignment improves the reference network's ability to comprehend complex scenes, capture meaningful object interactions, and effectively localize multiple objects under uncertain and challenging conditions. Through this process, the reference network benefits from the auxiliary network, enabling it to achieve enhanced semantic understanding directly from segmented point-cloud data.

To summarize, our contributions are as follows:
\begin{itemize}

    \item We introduce a hierarchical processing strategy that enables step-by-step object localization, progressively refining the identification of objects. This approach mimics human-like detection, allowing the model to handle complex scenes with varying object counts more effectively. 
    \item We employ a Contrastive Siamese Transformer framework consisting of two identical networks: one learning from ground-truth labels and the other operating on point-cloud data. The contrastive mechanism aligns their intermediate-layer outputs, enabling knowledge transfer from the auxiliary network to the reference network. This improves the model's understanding of complex scenes.
    \item Our model demonstrates superior performance in 3D grounding, excelling in scenarios with multiple objects. It not only outperforms existing models in multi-object tasks but also maintains strong performance in single-object tasks, making it versatile for various applications.
\end{itemize}

\section{Related Work}
\subsection{ 2D Language Grounding}
2D language grounding aims to locate a target object in a 2D image based on natural language descriptions. One-stage methods simplify this process by directly regressing target bounding boxes, integrating object detection and language understanding into a unified framework~\cite{liao2020real},~\cite{yang2019fast},~\cite{yang2020improving},\cite{wu2024image}. These methods prioritize efficiency by eliminating intermediate proposal generation, enabling end-to-end processing and joint language-vision alignment. In contrast, two-stage pipelines adopt a detection-and-selection approach: object proposals are first generated, and features for both image regions and language descriptions are extracted and aligned using attention mechanisms~\cite{deng2023transvg++}~\cite{li2021referring},~\cite{yang2019dynamic}. While offering higher accuracy through refined proposal-level feature extraction, two-stage methods usually incur greater computational costs. Recent works, such as TransVG~\cite{deng2023transvg++}, employ transformers to unify the strengths of both approaches, improving global context understanding and scalability in complex scenarios.

\subsection{3D Language Grounding}

3D language grounding extends vision-language alignment to 3D environments, where models localize objects in  3D point clouds based on natural language queries. Benchmarks like ScanRefer~\cite{chen2020scanrefer} and ReferIt3D~\cite{achlioptas2020referit3d} have driven research in this area. ScanRefer involves both detection and grounding, while ReferIt3D focuses on distinguishing target objects within the same semantic class. Many models integrate 2D and 3D data to enhance spatial reasoning~\cite{yang2021sat},~\cite{bakr2022look},~\cite{huang2022multi}, but they often struggle in cluttered or ambiguous scenes. Transformer-based methods~\cite{zhao2021transformer},~\cite{he2021transrefer3d},~\cite{yang2021sat} excel at capturing global context, making them effective for complex queries involving object attributes and spatial relations. However, most works focus on single-object grounding, limiting their applicability to real-world scenarios. Multi3DRefer~\cite{zhang2023multi3drefer} advances the field by tackling multi-object grounding, requiring models to localize multiple objects from a single query. Building on this, we propose H-COST, which leverages Hierarchical Processing and a Contrastive Siamese Transformer to improve multi-object grounding in complex and cluttered 3D environments.

\subsection{ Siamese Networks}

Siamese networks, as a foundational neural network architecture, have undergone significant evolution since their introduction by Bromley et al. for signature verification~\cite{bromley1993signature}. Designed to compare input pairs using shared-weight subnetworks for feature extraction and similarity computation, Siamese networks have since been applied across a wide range of tasks, including unsupervised~\cite{fahkr2017siamese}, semi-supervised~\cite{jia2021semi}, and self-supervised learning~\cite{li2021semantic}. Their flexibility and simplicity have made them a popular choice for tasks requiring pairwise comparison.

Siamese networks have been enhanced by integrating with architectures like Recurrent Neural Networks (RNNs)\cite{RNN}, Generative Adversarial Networks (GANs)\cite{goodfellow2014gan}, and Convolutional Neural Networks (CNNs)\cite{lecun1998cnn}. RNN-based Siamese networks leverage sequential modeling capabilities \cite{roy2022detection, xu2017siameserecurrent}, while GAN-based Siamese networks combine adversarial learning with similarity modeling \cite{tao2019resattrgan, zhang2020sienet}. CNN-based Siamese networks efficiently handle feature extraction and similarity comparison \cite{pei2016learning, tao2016siameseinstance}.

Over time, Siamese networks have been extended and refined to meet new challenges. For instance, Triplet Networks~\cite{hoffer2015deep} process three inputs simultaneously, using triplet loss to enhance the model’s ability to distinguish between similar and dissimilar pairs. Building on this concept, Quadruplet Networks~\cite{chen2017beyond} and DQAL~\cite{hou2019dual} extend the architecture further by introducing additional branches, enabling the simultaneous processing of multiple inputs. These multi-branch designs allow for more complex interactions among input samples, capturing richer relationships and improving discriminative power in tasks requiring fine-grained distinctions. In addition to multi-branch designs, Pseudo-Siamese Networks have emerged as a flexible adaptation of the original architecture. By allowing subnetworks to have different weights or structures, these networks are well-suited for tasks involving heterogeneous inputs or cross-modal matching~\cite{hughes2018identifying, treible2019wildcat}. Developments in contrastive learning have further propelled Siamese networks, with approaches like SimCLR~\cite{chen2020simple} and BYOL~\cite{grill2020bootstrap} demonstrating their effectiveness in unsupervised learning. These methods leverage augmented sample pairs to learn robust feature representations, highlighting the versatility of Siamese architectures.

Taking inspiration from Siamese network-based approaches, we propose H-COST for multi-object 3D grounding to address the challenge in 3D object localization tasks, where training models to directly localize objects from input 3D point clouds often lacks accurate semantic information. H-COST employs a contrastive Siamese framework with two identical networks—one capturing robust object relations from ground-truth labels and the other refining 3D point cloud representations—to enable more precise localization and semantic understanding in cluttered 3D scenes.

\section{PROPOSED METHOD}

\label{sec:pagestyle}
\begin{figure*}[hbt!]
  \centering
  \includegraphics[width=\textwidth]{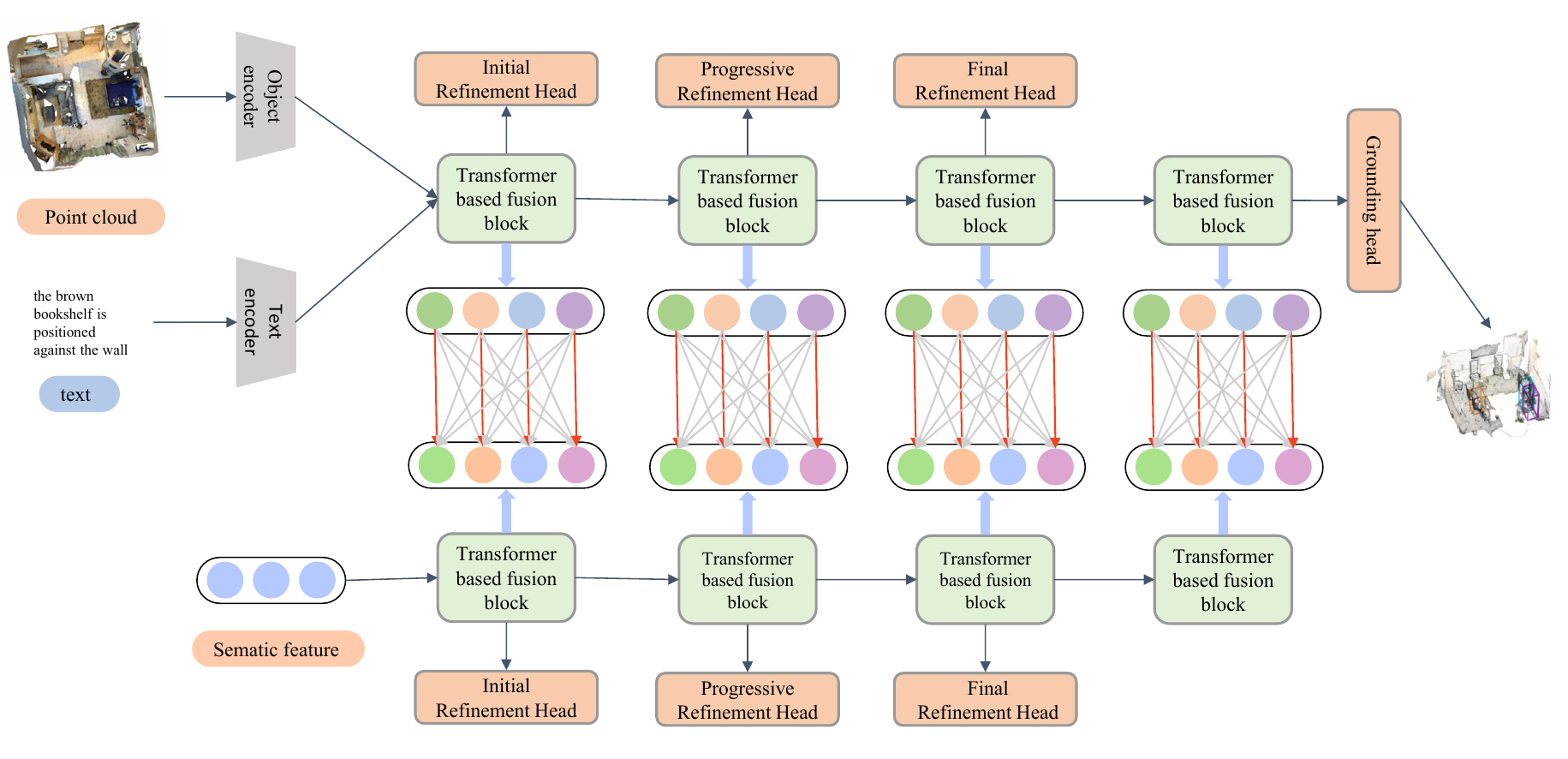} 
  \caption{\textbf{Overall architecture of H-COST.} This figure shows the integration of the auxiliary and inference networks. The auxiliary network processes ground-truth semantic features from text inputs, while the inference network extracts object features from raw point-cloud data. Both networks pass their respective features through transformer-based fusion blocks, which progressively refine the predictions using initial, progressive, and final refinement heads. The final grounded prediction is obtained through a grounding head that operates on the output of the final hidden state.}
  \label{fig:main}
\end{figure*}
\subsection{Overview}
Figure~\ref{fig:main} illustrates the overall architecture of H-COST. H-COST processes ground-truth object semantic features in the auxiliary network and raw point-cloud data in the inference network. Both networks utilize a text encoder to extract text features \( \textbf{F}_{\text{txt}} \in \mathbb{R}^{n_t \times d_t}\), while the inference network uses an object encoder to extract initial object features  \( \textbf{F}_{\text{obj}} \in \mathbb{R}^{n_o \times d_o}\), where \( n_o \) and \( n_t \) are the number of objects and text tokens, and \( d_o \) and \( d_t \) are their feature dimensions. These features are then fed into the transformer-based fusion block denoted as $\mathcal{B}_{\text{fusion}}$. In  $\mathcal{B}_{\text{fusion}}$, a hierarchical processing strategy is applied,  narrowing down the candidate object space by predicting objects within progressively smaller distance thresholds. Additionally, a contrastive learning mechanism aligns intermediate features and attentions between the auxiliary and inference networks by minimizing contrastive losses on attention maps \( \textbf{A}_{\text{aux}} \) and \( \mathbf{A}_{\text{inf}} \), as well as hidden states \( \textbf{H}_{\text{aux}} \) and \( \mathbf{H}_{\text{inf}} \). Finally, the prediction of the target object is obtained by applying a grounding head to the output of the final hidden state.

\subsection{ Transformer-based fusion blocks}

Both the auxiliary network and the inference network are composed of multiple transformer-based fusion blocks, denoted as $\mathcal{B}_{\text{fusion}}$. Each $\mathcal{B}_{\text{fusion}}$ contains a spatial-aware attention layer, a cross-attention layer, and a feed-forward layer.

In the attention mechanism, input features $\textbf{X}$ are first projected into queries ($\textbf{Q}$), keys ($\textbf{K}$), and values ($\textbf{V}$) using learned linear transformations:

\begin{equation}
\textbf{Q} = \textbf{W}_Q \mathbf{X}, \quad \textbf{K} =\textbf{ W}_K\textbf{ X}, \quad \textbf{V} = \textbf{W}_V \textbf{X},
\label{eq1}
\end{equation}

where $q$, $k$, and $v$ represent the object features. The standard attention scores are then computed as:

\begin{equation}
\text{Attn}(\textbf{Q}, \textbf{K}) = \frac{\textbf{Q}\textbf{K}^\top}{\sqrt{d_k}} .
\label{eq2}
\end{equation}

\subsubsection{ spatial-aware attention}
To incorporate spatial information, spatial weights \( \mathbf{h}_i^s \) are computed by applying a linear transformation to the sum of object features  \( \mathbf{F}_{obj} \in \mathbb{R}^{n_o \times d_o} \) and language embeddings \( \mathbf{F}_{txt} \in \mathbb{R}^{n_t \times d_t} \). These spatial weights are then combined with spatial features \( \mathbf{P}_{ij} \) via a dot product, followed by adding a bias and applying a sigmoid function: 

\begin{equation}
\textbf{A}_{\text{spatial}} = \sigma\left( \mathbf{h}_i^s \cdot \mathbf{P}_{ij} + \mathbf{b}_{\text{spatial}} \right).
\label{eq3}
\end{equation}

Here, the sigmoid function \( \sigma \) ensures that the output is normalized between 0 and 1, making it suitable for use as attention weights.
To integrate both spatial and standard attention, the spatial attention weights $A_{\text{spatial}}$ are fused with the standard attention scores through a log-sum combination:

{
 \begin{equation}
\text{SpatialAttn}(\textbf{Q},\textbf{ K}) = \text{softmax}\left( \log(\textbf{A}_{\text{spatial}}) + \frac{\textbf{Q}\textbf{K}^\top}{\sqrt{d_k}} \right).
\label{eq4}
\end{equation}
}

This approach allows the model to jointly reason about geometric relationships between objects, spatial configurations, and language cues, thereby enhancing its ability to infer object interactions and spatial arrangements.

\subsubsection{ cross-attention}

The cross-attention module further aligns object features \( \mathbf{F}_{\text{obj}} \) with text embeddings \( \mathbf{F}_{\text{txt}} \) to improve multimodal understanding. Specifically, object features are treated as queries \( \textbf{Q} \), while text embeddings are used as keys \( \textbf{K} \) and values \( \textbf{V} \). The cross-attention scores are computed as:

\begin{equation}
\textbf{Q} =\textbf{ W}_Q \mathbf{F}_{\text{obj}}, \quad \textbf{K} =\textbf{ W}_K \mathbf{F}_{\text{txt}}, \quad \textbf{V }= \textbf{W}_V \mathbf{F}_{\text{txt}}
\label{eq5}
\end{equation}

{
    \begin{equation}
    \text{Attn}_{\text{cross}}(\textbf{Q},\textbf{ K},\textbf{ V}) = \text{softmax}\left( \frac{\textbf{QK}^\top}{\sqrt{d_k}} \right)\textbf{V}.
    \label{eq6}
    \end{equation}
}

This mechanism enables the model to align object features with textual information, improving its ability to predict the relevance of each object to the text. The final output is generated by an MLP head, which predicts the relevance of each object to the text. A hierarchical processing strategy progressively refines these predictions across layers, as described in next section.

\subsubsection{Hierarchical Processing Strategy}

Previous methods primarily rely on a single-step prediction 
based on object proposals, which often lack the precision required for complex scenes. In contrast, we propose a hierarchical processing strategy that refines object localization, inspired by  human approach the task of identifying and locating objects. Just as a person first considers broader, easily distinguishable features to narrow down potential candidates, then gradually hones in on finer details for an accurate decision, our model progressively refines its predictions. This approach allows for more precise localization, addressing the limitations of single-step methods that struggle with densely packed objects or subtle spatial relationships.

At the end of each \( \mathcal{B}_{\text{fusion}} \) block, the model generates logits \( \mathbf{z}_i \) by applying a prediction head to the object embedding \( \mathbf{h}_i \) at layer \( i \), representing the likelihood of each object being the target. Based on these predicted logits, the objects are further refined at each stage by filtering them according to their spatial distance to the nearest target, using a distance threshold \( \delta_s \). The distance between object $i$ and the set of target objects $T$ is computed as the minimum distance to any target in $T$. Specifically, the distance \( \mathbf{d}_{iT} \) and the set of targets \( T_s \) for the current stage \( s \) are defined as follows:

\begin{equation}
d_{iT} = \min_{j \in T} \|\mathbf{c}_i - \textbf{c}_j\|_2 = \min_{j \in T} \sqrt{\sum_{k=1}^3 (\textbf{c}_{i,k} - \textbf{c}_{j,k})^2}.
\label{eq7}
\end{equation}
Objects whose distances to the nearest target are less than or equal to the threshold \( \delta_s \) are selected as the targets for that stage (\( T_s \)):

\begin{equation}
T_s = \{i : d_{iT} \leq \delta_s\}.
\label{eq8}
\end{equation}

We then compute the loss between the predicted logits and the selected targets \( T_s \). To balance the contribution of losses at different stages with other loss terms, we introduce a weighting factor \( \beta \) into the stage-wise loss. The loss for each stage is computed as:

\begin{equation}
\mathcal{L}_{hier} = \beta \cdot \frac{1}{|m|} \cdot \text{BCE}(\sigma(\textbf{z}_{n}), T_s),
\label{eq9}
\end{equation}

where BCE denotes the binary cross-entropy loss, \( \sigma(z_n) \) represents the sigmoid activation applied to the predicted logits, and \( T_s \) corresponds to the target labels derived from the distance threshold \( \delta_s \). Here, $m$ represents the number of valid samples used in the loss computation. The weighting factor \( \beta \) adjusts the overall contribution of hierarchical losses relative to other loss components in the model.

Thus, at each stage, only objects within the threshold distance are treated as targets, and the model optimizes its predictions by minimizing the weighted loss between the predicted logits and these threshold-based target objects. This formulation ensures that the model progressively refines its predictions based on spatial proximity, with each stage focusing on objects that are increasingly closer to the target, while \( \beta \) ensures flexibility in coordinating the hierarchical loss with other objectives in the overall optimization process.

\subsection{ Contrastive Siamese Transformer framework }

We define our Siamese network as consisting of an auxiliary network and an inference network, which share the same architecture but process different input data. Specifically, the auxiliary network takes ground-truth object semantic features as input, while the inference network operates on raw point-cloud data. This architecture allows both networks to learn comparable high-level representations from their respective inputs.

\subsubsection{Contrastive Loss in Siamese Networks}

The Siamese network is designed to learn a shared embedding space where the similarity between two input samples can be quantified. The training objective is based on contrastive loss. which can be defined as:

{\footnotesize
\begin{align}
\mathcal{L}_{\text{contrastive}} = 
\begin{cases} 
    \frac{1}{2} \| f(\textbf{x}_1) - f(\textbf{x}_2) \|_2^2, & \text{if } y = 1, \\[8pt]
    \frac{1}{2} \big(\max(0, m - \| f(\textbf{x}_1) - f(\textbf{x}_2) \|_2)\big)^2, & \text{if } y = 0,
\end{cases}
\end{align}
}

where \( f(x) \) denotes the embedding of the input sample \( x \), and \( \| f(x_1) - f(x_2) \|_2 \) is the Euclidean distance between the embeddings of two input samples \( x_1 \) and \( x_2 \). The binary label \( y \in \{0, 1\} \) indicates whether the two samples belong to the same class (\( y = 1 \)) or different classes (\( y = 0 \)). A fixed margin \( m > 0 \) is introduced to enforce separation between embeddings of different classes.

For samples from the same class (\( y = 1 \)), the loss minimizes the embedding distance:

\begin{equation}
\mathcal{L}_{\text{contrastive}} = \frac{1}{2} \| f(\textbf{x}_1) - f(\textbf{x}_2) \|_2^2.
\label{eq10}
\end{equation}

For samples from different classes (\( y = 0 \)), the loss ensures that the embedding distance exceeds the margin \( m \):

\begin{equation}
\mathcal{L}_{\text{contrastive}} = \frac{1}{2} \max(0, m - \| f(\textbf{x}_1) - f(\textbf{x}_2) \|_2)^2.
\label{eq11}
\end{equation}

This loss encourages the model to pull embeddings of similar samples closer together while pushing embeddings of dissimilar samples farther apart. Building upon this principle, our proposed framework extends the contrastive learning paradigm to align semantic representations between auxiliary and inference networks.

\subsubsection{Training Process}

The auxiliary network takes ground-truth object semantic features as input, denoted as \( \mathbf{X}_{\text{aux}} = \{x_{\text{aux}, i}\}_{i=1}^{N} \), where \( x_{\text{aux}, i} \) is the semantic feature of object \( i \). The network generates high-level representations \( \textbf{H}_{\text{aux}} = f_{\text{aux}}(\mathbf{X}_{\text{aux}}) \) and attention maps \( \textbf{A}_{\text{aux}} = g_{\text{aux}}(\mathbf{X}_{\text{aux}}) \). After training the auxiliary networks, the inference network, sharing the same structure, uses raw point-cloud data \( \mathbf{X}_{\text{inf}} = \{x_{\text{inf}, i}\}_{i=1}^{N} \) as input and computes representations \(\textbf{H}_{\text{inf}} = f_{\text{inf}}(\mathbf{X}_{\text{inf}}) \) and attention maps \( \textbf{A}_{\text{inf}} = g_{\text{inf}}(\mathbf{X}_{\text{inf}}) \), where \( f_{\text{inf}} \) and \( f_{\text{aux}} \) have identical architectures.

\subsubsection{Alignment Loss}

To ensure that the inference network learns similar attention maps and hidden states to the auxiliary networks, we employ alignment losses on both the attention maps and the hidden states. The goal is to minimize the distance between the corresponding attention maps and hidden states from the two networks for the same object. For each object \( i \), we compute the alignment losses as:

\begin{equation}
\mathcal{L}_{\text{align}}^{\textbf{A}} = \frac{1}{N} \sum_{i=1}^{N} \| \textbf{A}_{\text{inf}, i} - \textbf{A}_{\text{aux}, i} \|_2^2,
\label{eq12}
\end{equation}

\begin{equation}
\mathcal{L}_{\text{align}}^{\textbf{H}} = \frac{1}{N} \sum_{i=1}^{N} \| \textbf{H}_{\text{inf}, i} - \textbf{H}_{\text{aux}, i} \|_2^2.
\label{eq13}
\end{equation}

These alignment losses ensure that both the attention maps and hidden states learned by the inference network for object \( i \) are close to those learned by the auxiliary networks.

\subsubsection{Distinctiveness Loss}

To enforce distinctiveness between the hidden state representations of different objects, we introduce a distinctiveness loss. For each object \( i \) in the inference network, we compare its hidden state \( h_{\text{inf}, i} \) with the hidden states of all other objects \( j \neq i \) from the auxiliary networks. We enforce a margin \( m \), ensuring that the distance between these hidden states remains above a certain threshold. The distinctiveness loss is defined as:

{
\begin{equation}
\mathcal{L}_{\text{distinct}} = \frac{1}{N^2} \sum_{i=1}^{N} \sum_{j \neq i} \max(0, m - \| \textbf{H}_{\text{inf}, i} - \textbf{H}_{\text{aux}, j} \|_2)^2.
\label{eq14}
\end{equation}
}

Here, \( N \) is the number of objects, \( H_{\text{inf}, i} \) is the hidden state of object \( i \) from the inference network, and \( H_{\text{aux}, j} \) is the hidden state of object \( j \) from the auxiliary networks. The margin \( m \) ensures that the hidden states of different objects remain sufficiently distinct. This loss encourages the model to maintain a margin between the hidden state representations of different objects, ensuring that negative samples are not too similar, thus promoting better feature separation in the hidden space.

\subsubsection{Siamese Contrastive Loss}

To combine the alignment and distinctiveness objectives, we define the Siamese contrastive loss, which incorporates both the alignment losses (\( \mathcal{L}_{\text{align}}^{\textbf{A}} \) and \( \mathcal{L}_{\text{align}}^{\textbf{H}} \)) and the distinctiveness loss (\( \mathcal{L}_{\text{distinct}} \)). The Siamese contrastive loss is formulated as:

\begin{equation}
\mathcal{L}_{\text{siam\_contra}} = \alpha \cdot \mathcal{L}_{\text{distinct}} + \mathcal{L}_{\text{align}}^{\textbf{A}} + \mathcal{L}_{\text{align}}^{\textbf{H}},
\label{eq15}
\end{equation}

where \( \alpha \) is a weighting factor that balances the contribution of the distinctiveness loss and the alignment losses. This combined loss ensures that the hidden states and attention maps generated by the inference network match those generated by the auxiliary networks, as enforced by \( \mathcal{L}_{\text{align}}^{\textbf{A}} \) and \( \mathcal{L}_{\text{align}}^{\textbf{H}} \), while also ensuring that the hidden state representations for different objects are sufficiently separated, as enforced by \( \mathcal{L}_{\text{distinct}} \). By combining these objectives, the Siamese contrastive loss ensures that the inference network learns representations and attention maps that are both consistent with the auxiliary networks and discriminative across objects, helping the model achieve better generalization while maintaining feature distinctiveness in the latent space.

\section{Experiment}
This section presents the experimental evaluation of the proposed H-COST model on the Multi3DRefer\cite{zhang2023multi3drefer} dataset, which serves as a challenging benchmark for multi-object grounding in 3D scenes. We describe the experimental setup, dataset, comparison with other method and parameter tuning.

\subsection{experiment setting}
The experiments are conducted to evaluate the performance of H-COST and analyze the impact of its loss configurations.\textbf{ }The training process uses an initial learning rate of 0.0005 with the AdamW \cite{loshchilov2017adamw} optimizer. Both the auxiliary and reference networks are trained for 50 epochs, balancing efficiency and robustness. 

The experiments are conducted on the Multi3DRefer \cite{zhang2023multi3drefer} validation set which is based on ScanRefer\cite{chen2020scanrefer}, providing a solid foundation for evaluating the model's performance across five grounding tasks. The Multi3DRefer \cite{zhang2023multi3drefer} dataset is specifically designed for multi-object 3D grounding tasks. It provides a challenging benchmark for models tasked with localizing a variable number of objects based on natural language descriptions . Each scene consists of 3D point clouds with variable number of objects, and the language descriptions range from simple object references to complex spatial relationships. The dataset supports a variety of tasks, including single-object grounding, multi-object grounding, and zero-shot grounding, making it ideal for evaluating the versatility of grounding models. It is divided into training, validation, and test sets, each containing scenes of varying complexity to ensure comprehensive evaluation.

Performance is measured using the F1@0.5 metric. A prediction is considered correct if the Intersection over Union (IoU) between the predicted bounding box and the ground truth exceeds 50\%. This metric provides a robust evaluation of the model’s accuracy in grounding tasks, ensuring that only precise predictions are counted as correct. The use of F1@0.5 reflects the model’s ability to handle both precision and recall effectively, particularly in challenging scenarios involving distractors or multiple objects.

\subsection{Main Results}

Table \ref{overall} presents the comparative results of H-COST and the baseline methods on the validation set of the Multi3DRefer dataset. The results indicate that H-COST significantly outperforms the baseline models across multiple tasks, particularly in the multi-object grounding task.

\begin{figure*}[htbp]
    \centering
    \includegraphics[width=0.9\linewidth]{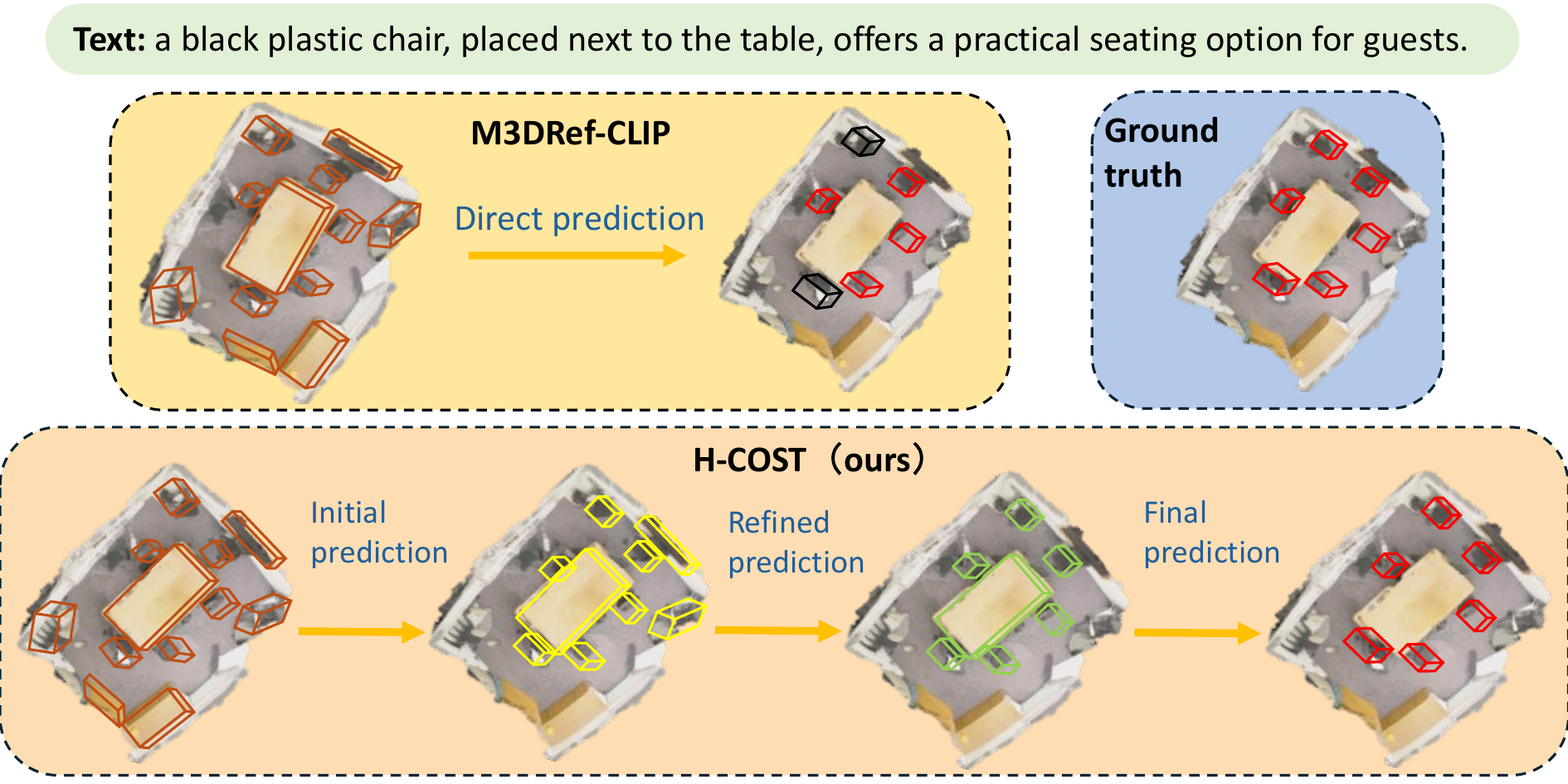}   \caption{ Qualitative results of M3DRef-CLIP versus H-COST on Multi3DRefer using predicted boxes. Brown boxes indicate object proposals, yellow boxes represent initially predicted objects, green boxes are refined predictions, red boxes are true positives with IoU threshold $\tau_{\text{pred}} > 0.5$, and black boxes denote missed objects.}
    \label{fig:qualitative}
\end{figure*}

\noindent\normalsize

\begin{table}[h!]
\centering
\caption{Comparison of F1@0.5 on the Multi3DRefer val set.}
\resizebox{\columnwidth}{!}{
\begin{tabular}{lcccccc}
\toprule
\multirow{2}{*}{\textbf{Method}} & \multicolumn{6}{c}{F1@0.5 on Val (\textuparrow)} \\
\cmidrule(lr){2-7}
 & \textbf{ZT w/o D} & \textbf{ZT w/D} & \textbf{ST w/o D} & \textbf{ST w/D} & \textbf{MT} & \textbf{All} \\
\midrule
3DVG-Trans \cite{zhao2021transformer}& 87.1 & 45.8 & 27.5 & 16.7 & 26.5 & 25.5 \\
D3Net\cite{chen2022d3net}& 81.6 & 32.5 & 38.6 & 23.3 & 35.0 & 32.2 \\
3DJCG\cite{cai2022_3djcg}& 94.1 & 66.9 & 26.0 & 16.7 & 26.2 & 26.6 \\
M3DRef-CLIP\cite{zhang2023multi3drefer}& 81.8 & 39.4 & 47.8 & 30.6 & 37.9 & 38.4 \\
\textbf{H-COST} & 78.88 & 41.49 & 61.10 & 39.45 & 48.39 & \textbf{47.94} \\
\bottomrule
\end{tabular}
}\label{overall}
\end{table}
In particular, H-COST shows strong performance in the multi-target grounding task, achieving an F1 score of \textbf{47.94}, \textbf{9.5\%} higher than the score of M3DRef-CLIP\cite{zhang2023multi3drefer}, the previously best-performing model. Figure~\ref{fig:qualitative} shows qualitative results

\subsection{Ablation Studies}

We conduct comprehensive ablation studies to analyze the impact of the distinctiveness loss weight ($\alpha$) and the hierarchical loss weight ($\beta$). The results, as shown in Table~\ref{tab:alpha_ablation} and \ref{tab:beta_ablation}, systematically explore these hyperparameters to achieve an optimal balance between the two losses.

\begin{table}[htbp]
\centering
\caption{F1@0.5 on Multi3DRefer validation set with different $\alpha$ configurations (fixed $\beta = 0$).}
\renewcommand{\arraystretch}{1.1} 
\setlength{\tabcolsep}{2.5pt} 
\resizebox{\linewidth}{!}{ 
\begin{tabular}{lccccccc}
\toprule
\textbf{Model} & \textbf{$\alpha$ }& \textbf{ST w/ D} & \textbf{ST w/o D} & \textbf{MT} & \textbf{ZT w/ D} & \textbf{ZT w/o D}& \textbf{All} \\
\midrule
H-COST& 0& 35.05& 47.8 & 37.9 & 39.4 & 81.8 & 38.4 \\
H-COST & 0.1 & 38.88 & 61.34 & 48.77 & 22.63 & 66.54 & 46.44 \\
H-COST & 0.2 & 39.53 & 60.57 & 48.29 & 34.11 & 77.67 & \textbf{47.46} \\
H-COST & 0.3 & 39.49 & 60.86 & 47.81 & 30.82 & 74.78 & 47.20 \\
H-COST & 0.4 & 38.99 & 60.84 & 47.90 & 27.87 & 73.35 & 46.78 \\
H-COST & 0.5 & 39.60 & 60.64 & 47.85 & 32.99 & 75.83 & 47.30 \\
\bottomrule
\end{tabular}}
\label{tab:alpha_ablation} 
\end{table}

As shown in table~\ref{tab:alpha_ablation}, the performance on the Multi3DRefer validation set for different configurations of \( \alpha \)  is summarized. The results indicate that increasing \( \alpha \), which controls the weight of the distinctiveness loss, initially improves performance, with the best overall F1 score of 47.46 achieved at \( \alpha = 0.2 \). However, as $\alpha$ exceeds 0.2, performance degrades due to overemphasis on the distinctiveness loss, which over-penalizes object embedding similarity and disrupts feature alignment between the reference and auxiliary networks.

The distinctiveness loss is designed to enhance the separability of features corresponding to different objects within the same scene, specifically between the reference network and the auxiliary network. \textbf{By enforcing a margin between the representations of different objects in these two networks, the distinctiveness loss helps the model minimize ambiguity when grounding objects in complex scenarios. }This is particularly critical for distinguishing between visually or spatially similar objects.

\begin{table}[htbp]
\centering
\caption{F1@0.5 on Multi3DRefer validation set with different $\beta$ configurations (fixed $\alpha = 0.2$).}
\renewcommand{\arraystretch}{1.1} 
\setlength{\tabcolsep}{2.5pt} 
\resizebox{\linewidth}{!}{ 
\centering

\begin{tabular}{lccccccc}
\toprule
\textbf{Model} & \textbf{$\beta$} & \textbf{ST w/ D} & \textbf{ST w/o D} & \textbf{MT} & \textbf{ZT w/ D} &\textbf{ ZT w/o D} & \textbf{All} \\
\midrule
H-COST& 0& 39.53& 60.57& 48.29& 34.11& 77.67& 47.46\\
H-COST & 0.1 & 39.45 & 61.10 & 48.39 & 41.49 & 78.88 & \textbf{47.94} \\
H-COST & 0.2 & 39.26 & 61.24 & 48.43 & 38.84 & 76.93 & 47.72 \\
H-COST & 0.3 & 39.69 & 60.56 & 48.04 & 40.96 & 79.02 & 47.85 \\
H-COST & 0.4 & 38.92& 60.82& 48.04& 47.49& 81.48& 47.81\\
H-COST & 0.5 & 38.40 & 60.15 & 48.23 & 39.36 & 77.95 & 47.16 \\
\bottomrule
\end{tabular}}
\label{tab:beta_ablation} 
\end{table}

Table~\ref{tab:beta_ablation} presents the performance for different configurations of \( \beta \) with \( \alpha \) fixed at 0.2. The hierarchical loss, controlled by \( \beta \), enables the model to refine its predictions progressively through a multi-stage reasoning process. This design is inspired by how humans analyze complex scenarios—from coarse, high-level understanding to fine-grained, detailed reasoning, allowing the model to focus its attention step-by-step and narrow down the object search space. The results indicate that the best performance is achieved when \( \beta = 0.1 \), with an overall F1@0.5 score of \textbf{47.94}. However, increasing \( \beta \) beyond 0.1 leads to slight performance degradation. Overemphasizing the hierarchical loss forces the model to rely too much on intermediate stages for localization, disrupting optimization and hindering the integration of global and local information. 
\textbf{These findings highlight that a moderate hierarchical loss weight effectively guides the multi-stage refinement process, ensuring meaningful contributions from each stage to the final prediction.}

\section{Conclusion}

In this work, we propose \textbf{H}ierarchical \textbf{Co}ntrastive Siamese \textbf{T}ransformers (H-COST) for 3D multi-object grounding, addressing the challenges of localizing multiple objects in cluttered 3D environments. Combining a hierarchical processing strategy with a contrastive Siamese framework, our method refines object localization step-by-step, mimicking human-like detection. The alignment of auxiliary and inference networks enhances semantic understanding, while distinctiveness loss improves feature separation. Experimental results on the Multi3DRefer\cite{zhang2023multi3drefer} dataset demonstrate that H-COST significantly outperforms state-of-the-art methods, particularly in complex multi-object scenarios. H-COST's robust performance across diverse tasks highlights its potential for real-world applications requiring precise 3D object localization and scene understanding.

\bibliographystyle{IEEEtran}
\bibliography{refs}

\begin{thebibliography}{10}
\providecommand{\url}[1]{#1}
\csname url@samestyle\endcsname
\providecommand{\newblock}{\relax}
\providecommand{\bibinfo}[2]{#2}
\providecommand{\BIBentrySTDinterwordspacing}{\spaceskip=0pt\relax}
\providecommand{\BIBentryALTinterwordstretchfactor}{4}
\providecommand{\BIBentryALTinterwordspacing}{\spaceskip=\fontdimen2\font plus
\BIBentryALTinterwordstretchfactor\fontdimen3\font minus \fontdimen4\font\relax}
\providecommand{\BIBforeignlanguage}[2]{{%
\expandafter\ifx\csname l@#1\endcsname\relax
\typeout{** WARNING: IEEEtran.bst: No hyphenation pattern has been}%
\typeout{** loaded for the language `#1'. Using the pattern for}%
\typeout{** the default language instead.}%
\else
\language=\csname l@#1\endcsname
\fi
#2}}
\providecommand{\BIBdecl}{\relax}
\BIBdecl

\bibitem{peng2023openscene}
S.~Peng, K.~Genova, C.~M. Jiang, A.~Tagliasacchi, M.~Pollefeys, and T.~Funkhouser, ``Openscene: 3d scene understanding with open vocabularies,'' in \emph{Proceedings of the IEEE Conference on Computer Vision and Pattern Recognition (CVPR)}, 2023.

\bibitem{vil3dref}
S.~Chen, P.-L. Guhur, M.~Tapaswi, C.~Schmid, and I.~Laptev, ``Language conditioned spatial relation reasoning for 3d object grounding,'' in \emph{Proceedings of the IEEE/CVF Conference on Computer Vision and Pattern Recognition (CVPR)}, 2022.

\bibitem{wang2023embodiedscan}
T.~Wang, X.~Mao, C.~Zhu, R.~Xu, R.~Lyu, P.~Li, X.~Chen, W.~Zhang, K.~Chen, T.~Xue, X.~Liu, C.~Lu, D.~Lin, and J.~Pang, ``Embodiedscan: A holistic multi-modal 3d perception suite towards embodied ai,'' in \emph{Proceedings of the IEEE/CVF Conference on Computer Vision and Pattern Recognition (CVPR)}, 2023.

\bibitem{saynav}
A.~Rajvanshi, K.~Sikka, X.~Lin, B.~Lee, H.-P. Chiu, and A.~Velasquez, ``Saynav: Grounding large language models for dynamic planning to navigation in new environments,'' in \emph{Proceedings of the IEEE/CVF Conference on Robotics and Automation (ICRA)}, 2024.

\bibitem{polaris}
T.~Wang, H.~Lin, J.~Yu, and Y.~Fu, ``Polaris: Open-ended interactive robotic manipulation via syn2real visual grounding and large language models,'' in \emph{Proceedings of the IEEE/RSJ International Conference on Intelligent Robots and Systems (IROS)}, 2024.

\bibitem{achlioptas2020referit3d}
P.~Achlioptas, A.~Abdelreheem, F.~Xia, M.~Elhoseiny, and L.~J. Guibas, ``Referit3d: Neural listeners for fine-grained 3d object identification in real-world scenes,'' in \emph{European Conference on Computer Vision (ECCV)}, 2020.

\bibitem{yang2021sat}
Z.~Yang, S.~Zhang, L.~Wang, and J.~Luo, ``Sat: 2d semantics assisted training for 3d visual grounding,'' in \emph{Proceedings of the IEEE/CVF International Conference on Computer Vision (ICCV)}, 2021, pp. 1856--1866.

\bibitem{zhang2023multi3drefer}
Y.~Zhang, Z.~Gong, and A.~X. Chang, ``Multi3drefer: Grounding text description to multiple 3d objects,'' in \emph{Proceedings of the IEEE/CVF International Conference on Computer Vision (ICCV)}, 2023.

\bibitem{liao2020real}
Y.~Liao, S.~Liu, G.~Li, F.~Wang, Y.~Chen, C.~Qian, and B.~Li, ``A real-time cross-modality correlation filtering method for referring expression comprehension,'' in \emph{Proceedings of the IEEE Conference on Computer Vision and Pattern Recognition (CVPR)}, 2020.

\bibitem{yang2019fast}
Z.~Yang, B.~Gong, L.~Wang, W.~Huang, D.~Yu, and J.~Luo, ``A fast and accurate one-stage approach to visual grounding,'' in \emph{Proceedings of the IEEE Conference on Computer Vision and Pattern Recognition (CVPR)}, 2019.

\bibitem{yang2020improving}
Z.~Yang, T.~Chen, L.~Wang, and J.~Luo, ``Improving one-stage visual grounding by recursive sub-query construction,'' in \emph{Proceedings of the European Conference on Computer Vision (ECCV)}, 2020.

\bibitem{wu2024image}
W.~Wu, X.~Qiu, S.~Song, Z.~Chen, X.~Huang, F.~Ma, and J.~Xiao, ``Image augmentation agent for weakly supervised semantic segmentation,'' \emph{arXiv preprint arXiv:2412.20439}, 2024.

\bibitem{deng2023transvg++}
J.~Deng, Z.~Yang, D.~Liu, T.~Chen, W.~Zhou, Y.~Zhang, H.~Li, and W.~Ouyang, ``Transvg++: End-to-end visual grounding with language conditioned vision transformer,'' \emph{IEEE Transactions on Pattern Analysis and Machine Intelligence}, 2023.

\bibitem{li2021referring}
M.~Li and L.~Sigal, ``Referring transformer: A one-step approach to multi-task visual grounding,'' in \emph{Advances in Neural Information Processing Systems (NeurIPS)}, 2021.

\bibitem{yang2019dynamic}
S.~Yang, G.~Li, and Y.~Yu, ``Dynamic graph attention for referring expression comprehension,'' in \emph{Proceedings of the IEEE International Conference on Computer Vision (ICCV)}, 2019.

\bibitem{chen2020scanrefer}
D.~Z. Chen, A.~X. Chang, and M.~Nie{\ss}ner, ``Scanrefer: 3d object localization in rgb-d scans using natural language,'' in \emph{European Conference on Computer Vision (ECCV)}, 2020.

\bibitem{bakr2022look}
E.~Bakr, Y.~Alsaedy, and M.~Elhoseiny, ``Look around and refer: 2d synthetic semantics knowledge distillation for 3d visual grounding,'' in \emph{Advances in Neural Information Processing Systems (NeurIPS)}, 2022.

\bibitem{huang2022multi}
S.~Huang, Y.~Chen, J.~Jia, and L.~Wang, ``Multi-view transformer for 3d visual grounding,'' in \emph{Proceedings of the IEEE/CVF Conference on Computer Vision and Pattern Recognition (CVPR)}, 2022.

\bibitem{zhao2021transformer}
L.~Zhao, D.~Cai, L.~Sheng, and D.~Xu, ``3dvg-transformer: Relation modeling for visual grounding on point clouds,'' in \emph{Proceedings of the IEEE/CVF International Conference on Computer Vision (ICCV)}, 2021.

\bibitem{he2021transrefer3d}
D.~He, Y.~Zhao, J.~Luo, T.~Hui, S.~Huang, A.~Zhang, and S.~Liu, ``Transrefer3d: Entity-and-relation aware transformer for fine-grained 3d visual grounding,'' in \emph{Proceedings of ACM Multimedia (ACM MM)}, 2021, pp. 2344--2352.

\bibitem{bromley1993signature}
J.~Bromley, I.~Guyon, Y.~LeCun, E.~S{\"a}ckinger, and R.~Shah, ``Signature verification using a "siamese" time delay neural network,'' in \emph{Advances in Neural Information Processing Systems (NeurIPS)}, 1993.

\bibitem{fahkr2017siamese}
M.~W. Fahkr, M.~M. Emara, and M.~B. Abdelhalim, ``Siamese-twin random projection neural network with bagging trees tuning for unsupervised binary image hashing,'' in \emph{Proceedings of the 5th International Symposium on Computational and Business Intelligence}, 2017, pp. 14--19.

\bibitem{jia2021semi}
X.~Jia \emph{et~al.}, ``Semi-supervised multi-view deep discriminant representation learning,'' \emph{IEEE Transactions on Pattern Analysis and Machine Intelligence}, vol.~43, no.~7, pp. 2496--2509, Jul. 2021.

\bibitem{li2021semantic}
W.~Li, H.~Chen, and Z.~Shi, ``Semantic segmentation of remote sensing images with self-supervised multitask representation learning,'' \emph{IEEE Journal of Selected Topics in Applied Earth Observations and Remote Sensing}, vol.~14, pp. 6438--6450, 2021.

\bibitem{RNN}
D.~E. Rumelhart, G.~E. Hinton, and R.~J. Williams, ``Learning representations by back-propagating errors,'' \emph{Nature}, vol. 323, no. 6088, pp. 533--536, 1986.

\bibitem{goodfellow2014gan}
I.~Goodfellow, J.~Pouget-Abadie, M.~Mirza, B.~Xu, D.~Warde-Farley, S.~Ozair, A.~Courville, and Y.~Bengio, ``Generative adversarial networks,'' in \emph{Advances in Neural Information Processing Systems (NeurIPS)}, vol.~27, 2014, pp. 2672--2680.

\bibitem{lecun1998cnn}
Y.~LeCun, L.~Bottou, Y.~Bengio, and P.~Haffner, ``Gradient-based learning applied to document recognition,'' \emph{Proceedings of the IEEE}, vol.~86, no.~11, pp. 2278--2324, 1998.

\bibitem{roy2022detection}
D.~Roy, T.~Ishizaka, C.~K. Mohan, and A.~Fukuda, ``Detection of collision-prone vehicle behavior at intersections using siamese interaction lstm,'' \emph{IEEE Transactions on Intelligent Transportation Systems}, vol.~23, no.~4, pp. 3137--3147, 2022.

\bibitem{xu2017siameserecurrent}
X.~Xu, B.~Ma, H.~Chang, and X.~Chen, ``Siamese recurrent architecture for visual tracking,'' in \emph{Proceedings of the IEEE International Conference on Image Processing (ICIP)}, 2017, pp. 1152--1156.

\bibitem{tao2019resattrgan}
R.~Tao, Z.~Li, R.~Tao, and B.~Li, ``Resattr-gan: Unpaired deep residual attributes learning for multi-domain face image translation,'' \emph{IEEE Access}, vol.~7, pp. 132\,594--132\,608, 2019.

\bibitem{zhang2020sienet}
X.~Zhang, F.~Chen, C.~Wang, M.~Tao, and G.-P. Jiang, ``Sienet: Siamese expansion network for image extrapolation,'' \emph{IEEE Signal Processing Letters}, vol.~27, pp. 1590--1594, 2020.

\bibitem{pei2016learning}
S.~Pei-Xia, L.~Hui-Ting, and L.~Tao, ``Learning discriminative cnn features and similarity metrics for image retrieval,'' in \emph{Proceedings of the IEEE International Conference on Signal Processing, Communications, and Computing (ICSPCC)}, 2016, pp. 1--5.

\bibitem{tao2016siameseinstance}
R.~Tao, E.~Gavves, and A.~W.~M. Smeulders, ``Siamese instance search for tracking,'' in \emph{Proceedings of the IEEE Conference on Computer Vision and Pattern Recognition (CVPR)}, 2016, pp. 1420--1429.

\bibitem{hoffer2015deep}
E.~Hoffer and N.~Ailon, ``Deep metric learning using triplet network,'' in \emph{International Workshop on Similarity-Based Pattern Recognition}.\hskip 1em plus 0.5em minus 0.4em\relax Springer, 2015, pp. 84--92.

\bibitem{chen2017beyond}
W.~Chen, X.~Chen, J.~Zhang, and K.~Huang, ``Beyond triplet loss: A deep quadruplet network for person re-identification,'' in \emph{Proceedings of the IEEE Conference on Computer Vision and Pattern Recognition}, 2017, pp. 403--412.

\bibitem{hou2019dual}
J.~Hou, H.~Zeng, J.~Zhu, J.~Hou, J.~Chen, and K.-K. Ma, ``Deep quadruplet appearance learning for vehicle re-identification,'' \emph{IEEE Transactions on Vehicular Technology}, vol.~68, no.~9, pp. 8512--8522, Sep. 2019.

\bibitem{hughes2018identifying}
L.~H. Hughes, M.~Schmitt, L.~Mou, Y.~Wang, and X.~X. Zhu, ``Identifying corresponding patches in sar and optical images with a pseudo-siamese cnn,'' \emph{IEEE Geoscience and Remote Sensing Letters}, vol.~15, no.~5, pp. 784--788, May 2018.

\bibitem{treible2019wildcat}
W.~Treible, P.~Saponaro, and C.~Kambhamettu, ``Wildcat: In-the-wild color-and-thermal patch comparison with deep residual pseudo-siamese networks,'' in \emph{Proceedings of the IEEE International Conference on Image Processing}, 2019, pp. 1307--1311.

\bibitem{chen2020simple}
T.~Chen, S.~Kornblith, M.~Norouzi, and G.~Hinton, ``A simple framework for contrastive learning of visual representations,'' in \emph{Proceedings of the International Conference on Machine Learning}.\hskip 1em plus 0.5em minus 0.4em\relax PMLR, 2020, pp. 1597--1607.

\bibitem{grill2020bootstrap}
J.-B. Grill, F.~Strub, F.~Altch{\'e}, C.~Tallec, P.~H. Richemond, E.~Buchatskaya, C.~Doersch, B.~A. Pires, Z.~D. Guo, M.~G. Azar, B.~Piot, K.~Kavukcuoglu, R.~Munos, and M.~Valko, ``Bootstrap your own latent: A new approach to self-supervised learning,'' in \emph{Advances in Neural Information Processing Systems}, vol.~33, 2020, pp. 21\,271--21\,284.

\bibitem{loshchilov2017adamw}
I.~Loshchilov and F.~Hutter, ``Fixing weight decay regularization in adam,'' in \emph{Proceedings of the International Conference on Learning Representations (ICLR)}, 2017.

\bibitem{chen2022d3net}
D.~Z. Chen, Q.~Wu, M.~Nie{\ss}ner, and A.~X. Chang, ``D3net: A unified speaker-listener architecture for 3d dense captioning and visual grounding,'' in \emph{European Conference on Computer Vision (ECCV)}, 2022.

\bibitem{cai2022_3djcg}
D.~Cai, L.~Zhao, J.~Zhang, L.~Sheng, and D.~Xu, ``3djcg: A unified framework for joint dense captioning and visual grounding on 3d point clouds,'' in \emph{Proceedings of the IEEE/CVF Conference on Computer Vision and Pattern Recognition (CVPR)}, 2022.

\end{thebibliography}
\end{document}